\documentclass[letterpaper, 10 pt, conference]{ieeeconf}  

\IEEEoverridecommandlockouts                              

\usepackage{graphics} 
\usepackage{epsfig} 
\usepackage{mathptmx} 
\usepackage{times} 
\usepackage{amsmath} 
\usepackage{amssymb}  
\usepackage{algorithm}  
\usepackage[hidelinks]{hyperref}
\usepackage{url}
\usepackage{graphicx}
\usepackage{booktabs}
\usepackage{array}
\usepackage{algpseudocode}
\usepackage{amsmath}
\usepackage{amsfonts}
\usepackage{amssymb}
\usepackage{nicefrac}
\usepackage{microtype}
\usepackage{xcolor}
\usepackage{multirow}
\usepackage{inconsolata}  
\newcommand{\methodName}{\textsc{Tracc}}
\newcommand{\citep}[1]{\cite{#1}}

\title{\LARGE \bf
\textsc{Track-and-Complete}: Learning Humanoid Skills\\from a Single Failed Human Video
}

\author{Sarmad Idrees$^{1}$ and Jongeun Choi$^{1,*}$\\[3pt]
{\normalsize Project page: \href{https://tracc-humanoid.github.io}{\textcolor{purple}{\texttt{https://tracc-humanoid.github.io}}}}
\thanks{$^{*}$Corresponding author.}%
\thanks{$^{1}$Sarmad Idrees and Jongeun Choi are with the School of Mechanical Engineering, Yonsei
University, Seoul 03722, Korea (e-mail: {\tt\small sarmad@yonsei.ac.kr}; {\tt\small jongeunchoi@yonsei.ac.kr}).}%
\thanks{This work was supported by the National Research Foundation of Korea (NRF) grant funded
by the Korea government (MSIT) (No.RS-2026-25505230), by the Institute of Information \& Communications Technology Planning \& Evaluation (IITP)
grant funded by the Korea government (MSIT) (No.RS-2026-25524880), and by the Korea Institute of Science and Technology (KIST) Institutional Program (No. 2E33801-25-015).}%
}

\IEEEaftertitletext{%
\vspace{0.4em}
\centerline{\includegraphics[width=\textwidth]{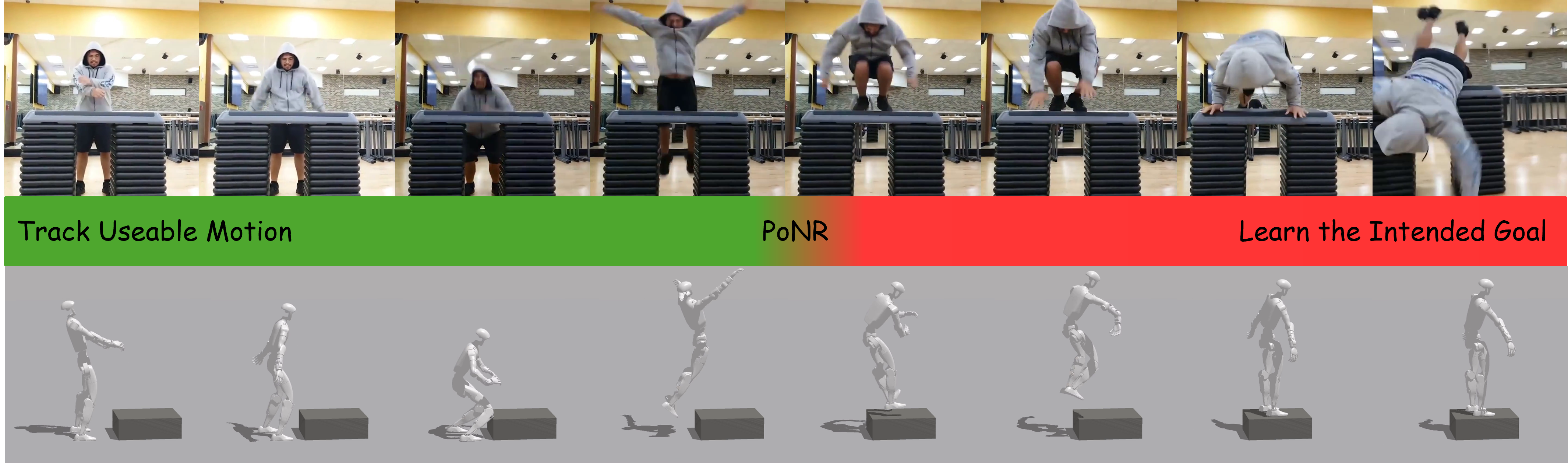}}
\vspace{0.4em}
\refstepcounter{figure}\label{fig:teaser}%
\noindent\parbox{\textwidth}{\footnotesize Fig.~\thefigure.\quad From a single failed video, \methodName~tracks the usable motion prefix (\textcolor{green}{green}) up to the Point-of-No-Return (PoNR), then optimizes the intended task outcome (\textcolor{red}{red}) without a successful task-motion demonstration.}
\vspace{0.8em}
}

\begin{document}

\maketitle
\thispagestyle{empty}
\pagestyle{empty}

\begin{abstract}
Learning humanoid skills from videos typically requires a successful human demonstration, which often demands custom data collection. 
Although failures have traditionally been treated only as negative examples in robot learning, they can still reveal a usable trajectory prefix before the task fails, as well as the intended outcome. 
To leverage this information from a failed-attempt video, we propose \methodName, a pipeline that imitates the useful portion of the motion trajectory and then completes the task based on the inferred task outcome. 
The usable motion prefix serves as prior knowledge until the failure occurs, after which the task-completion reward guides the policy to learn the intended task goal without requiring a successful task trajectory.
We evaluate our method on six in-the-wild failed human tasks from the Oops! dataset. 
Our experimental results demonstrate the effectiveness of the proposed approach for learning from failed attempts when no successful demonstration is available. 
Thus, these findings establish failed human videos as a viable source of supervision for humanoid skill learning.
\end{abstract}


\begin{table}[t]
  \centering
  \caption{Comparison of demonstration-based policy learning methods.}
  \label{tab:prior_work_comparison}
  \setlength{\tabcolsep}{1.5pt}
  \renewcommand{\arraystretch}{1.2}
  \scriptsize
  \begin{tabular}{@{}
      >{\raggedright\arraybackslash}p{0.31\columnwidth}
      >{\centering\arraybackslash}p{0.145\columnwidth}
      >{\centering\arraybackslash}p{0.18\columnwidth}
      >{\centering\arraybackslash}p{0.105\columnwidth}
      >{\centering\arraybackslash}p{0.17\columnwidth}@{}}
    \toprule
    Method & Embodiment & Demonstration & Single video & Use of failure \\
    \midrule
    VideoMimic \cite{allshire2025videomimic}   & Humanoid & Target video  & Multi & -- \\
    MeshMimic \cite{zhang2026meshmimic}        & Humanoid & Target video  & $\checkmark$    & -- \\
    HDMI \cite{weng2025hdmi}                   & Humanoid & Target video  & $\checkmark$ & -- \\
    OKAMI \cite{okami2024}                     & Humanoid & Target video  & $\checkmark$ & -- \\
    ORION \cite{zhu2026vision}                 & Arm      & Target video  & $\checkmark$ & -- \\
    LUCID \cite{gupta2026lucid}                & Arm      & Video corpus  & Multi & -- \\
    \midrule
    Donut as I Do \cite{grollman2011donut}     & Arm      & Failed motion & -- & Negative \\
    RL-VLM-F \cite{wang2024}                   & Arm      & Obs. pairs   & -- & Negative \\
    Goals from failure \cite{epstein2021learning} & None  & Failed videos & -- & Infer goal \\
    \midrule
    \methodName~(ours)                         & Humanoid & Failed video  & $\checkmark$ & Prefix + goal \\
    \bottomrule
    \addlinespace[2pt]
  \end{tabular}
\end{table}

\begin{figure*}[t]
  \centering
  \includegraphics[width=0.97\textwidth]{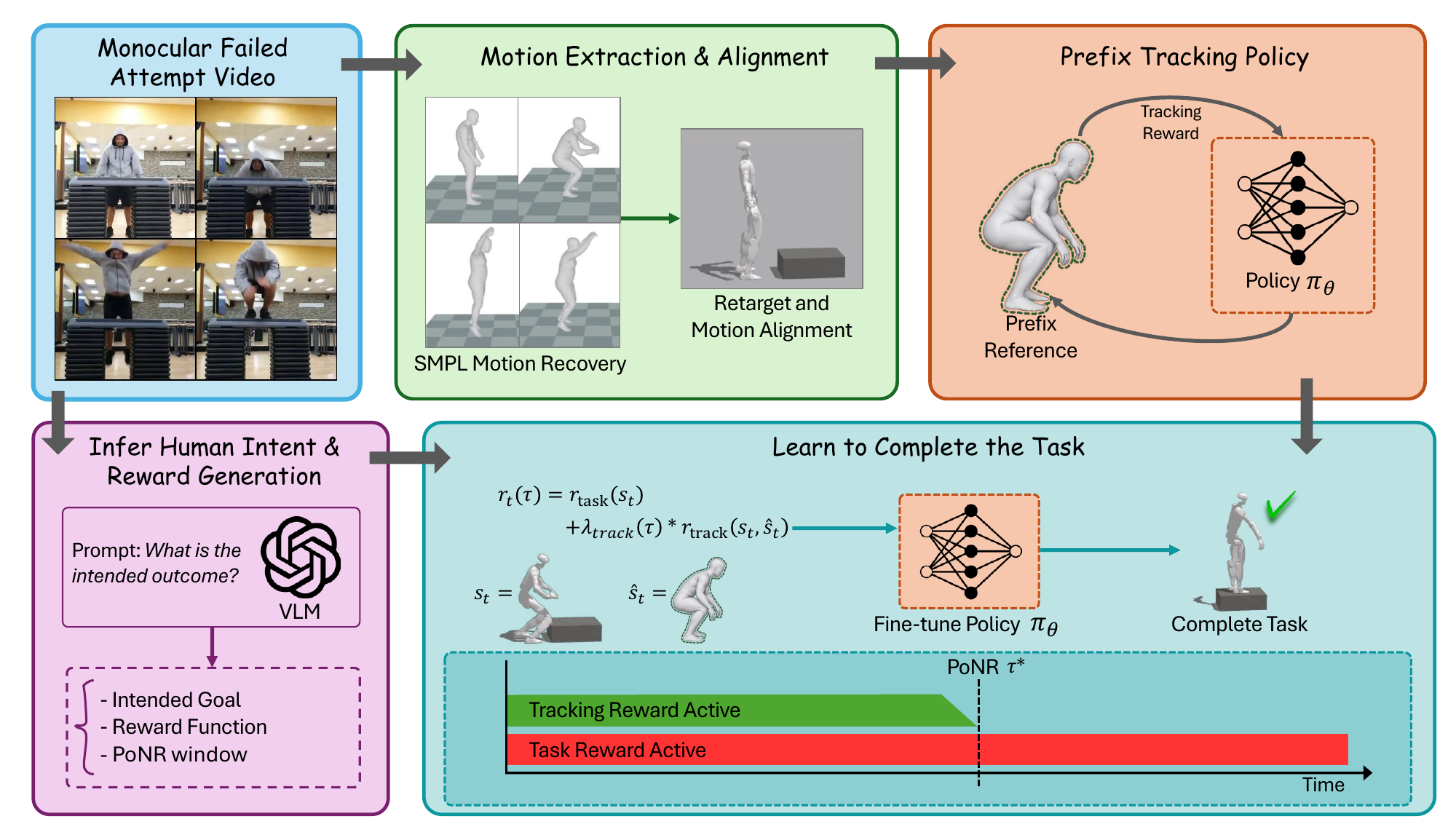}
  \caption{The overall architecture of the \methodName~pipeline. Human motion is reconstructed from the failed video, retargeted to the humanoid, and aligned with the simulated scene to train the prefix tracker. A VLM infers the intended goal, proposes task-reward candidates, and estimates a PoNR window. The tracker is then fine-tuned to complete the intended task.
  }
  \label{fig:main}
\end{figure*}

\section{Introduction}
\label{sec:intro}

Traditionally, learning physics-based humanoid skills from videos has focused on imitating demonstrated successful actions~\cite{allshire2025videomimic,weng2025hdmi,peng2018sfv, luo2023perpetual,he2024h2o}.
The recovered motion from these videos is therefore treated as a reference trajectory, and the policy is trained to reproduce the motion for a successful task completion. 
However, relying only on successful attempts often requires custom data collection, which is a tedious and expensive process.

On the other hand, failed attempts were mainly treated as negative examples that the policy should learn to avoid~\cite{grollman2011donut, wang2024}. 
A failed attempt can still show a useful approach, preparation, or early interaction before the
task goes wrong. 
It can also reveal what the person intended to achieve for the task completion. 
Treating the whole video as an imitation target also includes the failed ending, while treating the whole video only as a negative example discards its useful prefix.
Therefore, rather than treating a failed attempt entirely as either an imitation target or a negative example, we utilize its usable trajectory as motion guidance and its intended goal for task completion.

Using widely available failed attempts of human actions~\cite{epstein2020oops}, we study whether a humanoid can learn a complete task from this partial information when no successful task motion is available.
We propose \textbf{\methodName}, a pipeline that \underline{\textbf{Trac}}ks and \underline{\textbf{C}}omplete the task by first imitating usable motion prefix from unsuccessful demonstrations and then learning to achieve the intended outcome. 
Specifically, we divide a failed video into two forms of supervision: the usable trajectory prefix provides reference motion for how the task begins, while the inferred intent describes how the task should end.
The human motion is recovered from the video and retargeted to a humanoid robot, and the intendedtask outcome is inferred through a vision-language model (VLM), which also proposes task-completion reward and estimates when reference guidance should be released.

To determine how much of the failed demonstration should be used as reference motion, we define the Point-of-No-Return (PoNR) as the last point in the trajectory where the demonstrated motion remains useful for completing the task, before the observed failure begins.
Tracking reward fades as the policy approaches PoNR, then the task-completion reward takes control to guide the policy toward the inferred intended outcome (see Figure~\ref{fig:teaser}).
Thus, the method utilizes the usable portion of the failed attempt without imitating its failed ending or requiring a successful completion trajectory.
To the best of our knowledge, \methodName~is the first method to learn whole-body humanoid skills from a single failed human video, using the pre-failure motion as positive guidance and an inferred outcome for completion.

We evaluate \methodName~on six failed-action videos from the Oops! dataset~\cite{epstein2020oops}. 
The failed attempt videos are selected under the assumptions that at least some useful human motion prefix is available and that the intended final outcome is feasible in the simulation environment. 
The experiments show that our method successfully learns to track the initial guidance motion and subsequently complete the task when no successful task-completion trajectory is available.

The main contributions of this work are threefold:
\begin{enumerate}
    \item We formulate a video-based humanoid learning problem that learns a complete task from a failed human demonstration by exploiting its usable motion prefix and intended outcome, without requiring any successful task trajectory.
    \item We propose \methodName, which decomposes a failed demonstration into motion guidance and task intent, and combines retargeted human motion with a task reward to learn the intended task beyond the observed failure.
    \item We introduce the Point-of-No-Return (PoNR) to identify the latest viable part of the failed trajectory and progressively transition the policy from motion tracking to task-completion reward.
\end{enumerate}

\section{Related Work}
\label{sec:related}

\paragraph{Video-Based Humanoid Learning}
Physics-based imitation learning methods learn to mimic reference human motion~\cite{mahmood2019amass} by tracking rewards and reference-state initialization~\cite{peng2018deepmimic}, or learn motion style through an adversarial motion prior~\cite{peng2021amp,peng2022ase,tessler2023calm}.
Video-based methods first recover this reference motion from a monocular video and then train the policy to imitate the demonstrated task action~\cite{peng2018sfv,shen2024gvhmr,luo2023perpetual}.
Later works transferred the retargeted motion to humanoid robots for learning from human video demonstrations~\cite{he2024h2o,allshire2025videomimic,li2025robomirror,zhang2026meshmimic}.
These methods require a successful demonstration of the task as a behavior worth reproducing.
In this study, we argue that failed demonstrations also provide enough information to learn the intended task by utilizing only the part of a motion that remains useful before the observed failure.

\paragraph{Learning From Failed Demonstrations}
Prior work has used failed or imperfect demonstrations in several ways. Failed examples can identify regions that a policy should avoid~\cite{grollman2011donut}, confidence scores can reduce the influence of poor demonstrations~\cite{wu2019imperfect}, and rankings over suboptimal behavior can support reward inference~\cite{brown2019trex}. 
Failed videos have also been used to infer goals for visual planning~\cite{epstein2021learning}.
These approaches use failure as negative example, ranking, or only goal inference. However, learning a physics-based humanoid policy from a single failed monocular human video has not been explored. 
\methodName~uses the pre-failure motion as positive trajectory information and the inferred goal to complete the task.

\paragraph{Language Models for Reward Generation}
Language models have demonstrated major success in translating task descriptions into executable reward functions for
reinforcement learning \cite{yu2023language,xie2023text2reward,ma2024eureka}, and
related work extends reward generation for sim-to-real transfer~\cite{ma2024dreureka}. Video2Reward~\cite{zeng2024video2reward} generates imitation rewards from videos that
show target behavior. Vision-language models have also been used directly as reward models or preference providers \cite{rocamonde2023vlmrm,wang2024}. 
Building on these advancements, we use a VLM to infer the intended final outcome from the failed action and generate a task-completion reward that guides the policy after the usable motion ends.


\section{\methodName: Track-and-Complete}
\label{sec:method}

The overall pipeline is illustrated in Figure~\ref{fig:main}.
We treat a failed-attempt video sequence as a partial demonstration rather than a negative example.
First, the human motion is recovered and aligned with respect to the simulation environment, while a VLM generates a reward function after inferring the intended outcome of the task.
Second, a tracking policy is trained to imitate the motion up to the point where the failure begins to appear.
Finally, we further fine-tune the policy to first follow the useful trajectory and then complete the intended outcome.
Hereafter, we refer to this combined tracking and task-objective policy as a task-specific unified policy, where a separate policy is trained for each task.

\subsection{Problem Formulation}
Given a failed attempt human video $V=(I_1,\ldots,I_T)$, we aim to learn a policy $\pi_\theta$ that reproduces the useful trajectory of the observed motion and completes the intended task. 
The retargeted motion reference is defined as $\hat M=(\hat s_1,\ldots,\hat s_T)$, while the VLM produces an intended goal description, task-reward candidates $\{r_{\mathrm{task}}^k\}_{k=1}^{K}$, and a PoNR window $PoNR_w=[\tau_{\min},\tau_{\max}]$. 
At time $t$, the policy observes the humanoid states, the task observations, and prefix reference trajectory, and outputs normalized joint-position targets.

The unified policy is trained by uniformly sampling a release time $\tau$ per episode from $PoNR_w$ and maximizing
\begin{equation}
\theta^* = \arg\max_\theta \,
\mathbb{E}_{\tau\sim\mathcal{U}[\tau_{\min},\tau_{\max}],\,\pi_\theta}
\left[\sum_{t=0}^{H-1}\gamma^{ t} r_t(\tau)\right],
\label{eq:objective}
\end{equation}
where $\theta$ denotes the trainable policy parameters, $\theta^*$ their optimized values, and $\pi_\theta$ the resulting policy. Also, $H$ is the episode horizon, $\gamma\in[0,1)$ is the discount factor, and $r_t(\tau)$ is the reward at time $t$ under release time $\tau$.

\begin{algorithm}[t]
  \caption{\methodName~: Track-and-Complete}
  \label{alg:tracc}
  \scriptsize
  \begin{algorithmic}[1]
    \Require Failed-attempt video demonstration $V$
    \Ensure Unified policy $\pi_\theta$, selected reward $r_{\mathrm{task}}^{*}$, and measured PoNR $\tau^*$
    \Statex \textbf{Construct motion and task supervision}
    \State $\hat M \gets \Call{ReconstructRetargetAlign}{V,\mathcal{E}}$
    \State $(\{r_{\mathrm{task}}^k\}_{k=1}^{K},PoNR_w) \gets \Call{InferIntentAndRewards}{V}$
    \Statex \textbf{Learn the reference prefix}
    \State $\theta_{\mathrm{pre}} \gets \Call{TrainTracker}{\hat M,\text{uniform RSI}}$
    \State $\theta_{\mathrm{pre}} \gets \Call{FineTuneFromStart}{\theta_{\mathrm{pre}},\hat s_1}$
    \Statex \textbf{Select a task reward}
    \State $k^*\gets\Call{BestBySuccessRate}{\{r_{\mathrm{task}}^k\}_{k=1}^{K}}$; \quad $\theta\gets\theta_{\mathrm{pre}}$
    \Statex \textbf{Train one release-conditioned policy}
    \While{not converged}
      \ForAll{parallel environments $e$}
        \State Reset to $\hat s_1$ and sample $\tau_e\sim\mathcal{U}(PoNR_w)$
        \State Roll out $\pi_\theta$ with time-to-release in the observation
        \If{$t < \tau_e$}
          \State Use task reward and weighted tracking reward
        \Else
          \State Mask the reference and use task reward only
        \EndIf
      \EndFor
      \State Update $\theta$ with PPO using Eq.~\ref{eq:unified_reward}
    \EndWhile
    \State Measure the PoNR $\tau^*$ from terminal success
    \State \Return $\pi_\theta,r_{\mathrm{task}}^{k^*},\tau^*$
  \end{algorithmic}
\end{algorithm}

\begin{table}[t]
  \centering
  \caption{Policy observation space.}
  \label{tab:observation_space}
  \setlength{\tabcolsep}{2pt}
  \renewcommand{\arraystretch}{1.15}
  \scriptsize
  \begin{tabular}{@{}
      >{\raggedright\arraybackslash}p{0.20\columnwidth}
      >{\raggedright\arraybackslash}p{0.45\columnwidth}
      >{\centering\arraybackslash}p{0.10\columnwidth}
      >{\centering\arraybackslash}p{0.18\columnwidth}@{}}
    \toprule
    Group & Component & Dim. & After $\tau$ \\
    \midrule
    Proprioception & Root position, orientation, velocity & 13 & Observed \\
                   & Joint position   & 29 & Observed \\
                   & Joint velocity   & 29 & Observed \\
                   & Previous action  & 29 & Observed \\
    \midrule
    Task state & Task object, goal, and scene state & $d_{\mathrm{task}}$ & Observed \\
    \midrule
    Reference & Activity, phase, release time $\tau$ & 3 & Masked-out \\
              & Heading-relative root offset & 3 & Masked-out \\
              & Heading-relative hand and foot offsets & 12 & Masked-out \\
    \bottomrule
  \end{tabular}
\end{table}

\begin{figure*}[t]
  \centering
  \includegraphics[width=0.325\textwidth]{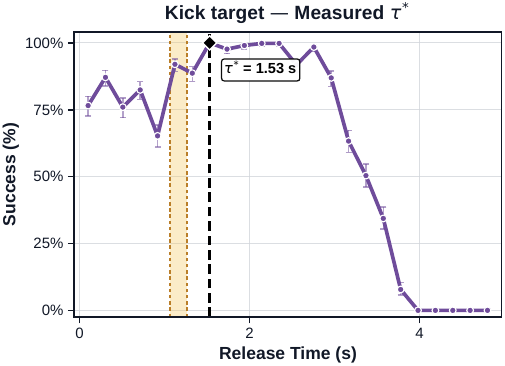}\hfill
  \includegraphics[width=0.325\textwidth]{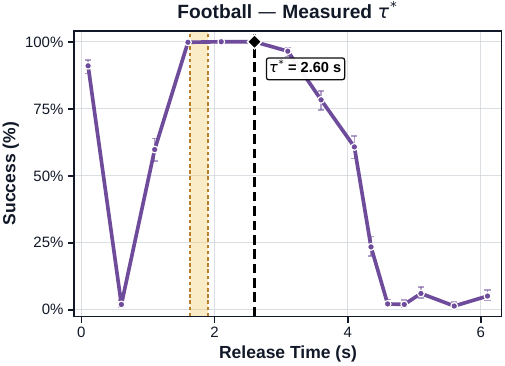}\hfill
  \includegraphics[width=0.325\textwidth]{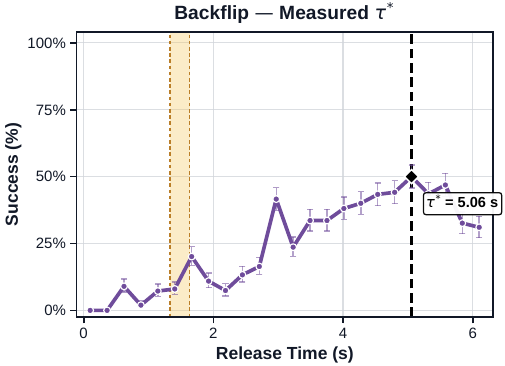}\\[4pt]
  \includegraphics[width=0.325\textwidth]{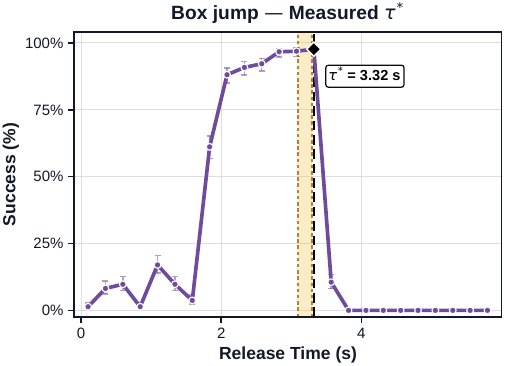}\hfill
  \includegraphics[width=0.325\textwidth]{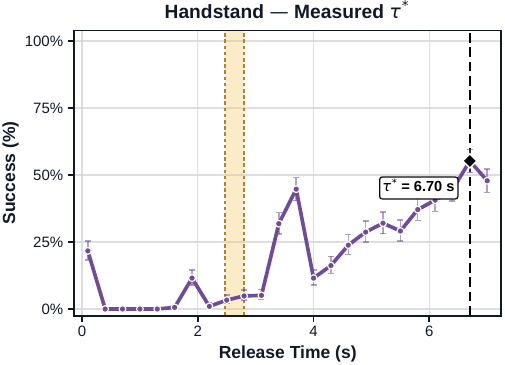}\hfill
  \includegraphics[width=0.325\textwidth]{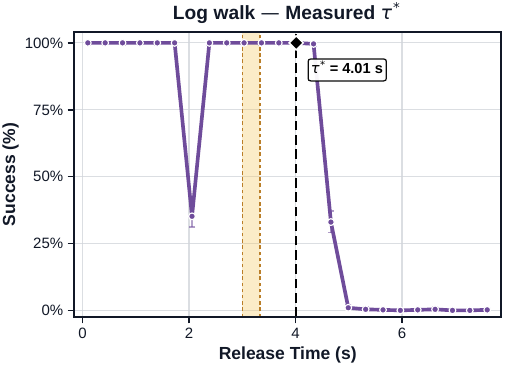}
  \caption{Task success vs. release time $\tau$ plots for each task. The shaded area denotes the VLM-proposed release window $PoNR_w$, and the dashed line marks the measured PoNR $\tau^*$.}
  \label{fig:ponr_curves}
\end{figure*}

\subsection{Motion Reconstruction}
We process the failed-attempt video in three steps. We first utilize GVHMR~\cite{shen2024gvhmr} to recover a world-grounded 3D human trajectory from the monocular video. 
Afterwards, we retarget the reconstructed body motion to the 29-DoF Unitree G1 while preserving its root motion and whole-body configuration within the humanoid embodiment. 
The retargeted motion is then aligned with respect to the simulated task scene. 
Furthermore, we estimate the task-object trajectory from the video and use the corresponding scene geometry to place the humanoid reference consistently with the object. The resultant scene-aligned reference motion $\hat M$ is finally used by the tracking stage.

\subsection{Prefix Reference Tracking}
From the provided video $V$, we only utilize the usable prefix of $\hat M$ as a positive imitation target. Where, the usable reference cutoff is determined by the VLM estimated $PoNR_w$. 
Following the motion-tracking objective of DeepMimic~\cite{peng2018deepmimic}, we define the tracking reward as a weighted sum of pose ($q$), joint-velocity ($\dot q$), end-effector ($\mathrm{ee}$), and root ($\mathrm{root}$) tracking terms:
\begin{equation}
r_{\mathrm{track}}(s_t,\hat{s}_t)
= \frac{1}{\sum_{m\in\mathcal{M}}w_m}
\sum_{m\in\mathcal{M}} w_m\exp(-k_m e_m),
\label{eq:tracking_reward}
\end{equation}
with $\mathcal{M}=\{q,\dot q,\mathrm{ee},\mathrm{root}\}$, where $s_t$ and $\hat s_t$ are the current humanoid states and reference states, $e_m$ is the corresponding mean-squared tracking error, and $w_m$ and $k_m$ control the contribution and sensitivity of each term.

We first train the tracking policy with reference-state initialization (RSI), sampling the humanoid reset state uniformly along the reference motion trajectory. RSI exposes the policy directly to every phase of the motion instead of requiring an untrained controller to reach later reference states through a long rollout, a strategy established for physics-based motion tracking~\cite{peng2018deepmimic}.
However, RSI alone can hide errors that accumulate when the motion is executed from the initial state. 
As our final goal is to learn a task-conditioned policy that must reliably track the usable trajectory from the initial state and then continue to complete the intended task after the usable trajectory ends, we further fine-tune the same tracker with resets restricted to the initial reference frame. This reset condition forces the policy to reach later prefix states through its own dynamics and matches the initial-state distribution used during the unified policy training.

\subsection{Intent Inference and Reward Generation}
Building on prior work that shows language models can translate semantic task descriptions into executable reinforcement-learning rewards~\cite{yu2023language,xie2023text2reward,ma2024eureka}, the VLM is prompted with the frame samples from the $V$ to infer the intended outcome, generate $K=3$ reward candidates for task-completion, and provide $PoNR_w$. 
Since the visual observation can work as the guidance, hence, we purposely request an interval (i.e. $PoNR_w$) rather than a single PoNR timestamp to preciesely predict when the demonstrated behavior begins to deteriorate.

During tracking and unified policy training, we sample release times $\tau$ uniformly from $PoNR_w$. 
This sampled $\tau$ acts as the boundary where the tracking reward should be faded out to learn the task completion instead of tracking the failure motion.
After training, we evaluate the same policy at several fixed release times over the full video timeframe and define the measured PoNR $\tau^*$ as the latest tested time that produces the best terminal success rate. 
Thus, the VLM-proposed $PoNR_w$ serves as a visual training prior, whereas the final $\tau^*$ is measured from the trained policy's success curve and reported in Sec.~\ref{sec:experiments}.

\begin{table*}[t]
  \centering
  \caption{Selected task rewards for unified policy training.}
  \label{tab:selected_rewards}
  \setlength{\tabcolsep}{4pt}
  \renewcommand{\arraystretch}{1.12}
  \scriptsize
  \begin{tabular}{@{}
      >{\raggedright\arraybackslash}p{0.07\textwidth}
      >{\raggedright\arraybackslash}p{0.9\textwidth}@{}}
    \toprule
    Task & Selected reward terms and weights \\
    \midrule
    Kick target & Pad approach ($0.10$); latched strike ($0.30$); $\times$ after the strike: two-foot contact ($0.22$), pelvis height ($0.18$), and stability and uprightness ($0.15$); not fallen ($0.05$). \\
    Football & Foot--ball approach ($0.12$); ball contact ($0.30$); ball speed toward the goal ($0.22$); ball--goal distance ($0.18$); uprightness ($0.14$); not fallen ($0.04$); hovering without contact penalty ($-0.10$). \\
    Backflip & Landing-region approach ($0.08$); left and right foot contact with load ($0.15$ each); landing pelvis height ($0.18$); low linear ($0.18$) and angular ($0.12$) velocity; uprightness ($0.09$); latched inversion bonus ($0.05$). \\
    Box jump & Feet near the box top ($0.15$); loaded top contact ($0.30$); standing height ($0.25$); pelvis over the box ($0.10$); controlled standing ($0.10$); uprightness ($0.05$); settling ($0.05$); non-top ($-0.55$) and invalid-contact ($-0.10$) penalties; reward set to $1$ on success. \\
    Handstand & Hand-patch approach ($0.12$); left and right palm contact ($0.15$ each); bilateral palm contact ($0.08$); $\times$ after bilateral contact: inversion progress ($0.18$) and feet above the pelvis ($0.22$); pelvis height ($0.18$); settling ($0.07$). \\
    Log walk & Progress toward the far end of the log ($0.45$); foot contact on the log top ($0.30$); pelvis height above the log ($0.15$); uprightness ($0.10$); hovering without contact penalty ($-0.20$). \\
    \bottomrule
  \end{tabular}
\end{table*}



\subsection{Unified Policy Training}
\label{sec:unified_training}
The unified policy is initialized with the prefix tracker policy and further fine-tuned to perform a full reference tracking and task-completion from the initial reset state. 
Each VLM reward proposal $r_{\mathrm{task}}^k$ is trained with same iteration budget, and best overall success reward proposal is selected for further unified policy training. Table~\ref{tab:selected_rewards} summarizes the selected task reward for each task.



To achieve unified tracking and task-completion objective training, we propose a combined reward function, where the task reward remains active throughout the episode, while the tracking reward guides the reference tracking and fades before the sampled release time $\tau$:
\begin{equation}
r_t(\tau)=r_{\mathrm{task}}(s_t)+\lambda_{track}(\tau)r_{\mathrm{track}}(s_t,\hat{s}_t),
\label{eq:unified_reward}
\end{equation}
where $\lambda_{track}$ is set to $0.5$ at the beginning of each episode, and decreases linearly to zero over the final 10\% before $\tau$, and remains zero thereafter.

To facilitate this transition from tracking to task-completion learning, we provide release time $\tau$ as an observation to the policy. 
Therefore, the policy can anticipate when reference guidance will disappear instead of reacting to an unexpected observation shift. 
At and after $\tau$, the entire reference trajectory block is masked to zero, and the policy is required to complete the task from robot states and task observations alone. Algorithm~\ref{alg:tracc} summarizes the whole \methodName~pipeline.

\begin{figure}[t]
  \centering
  \includegraphics[width=0.97\linewidth]{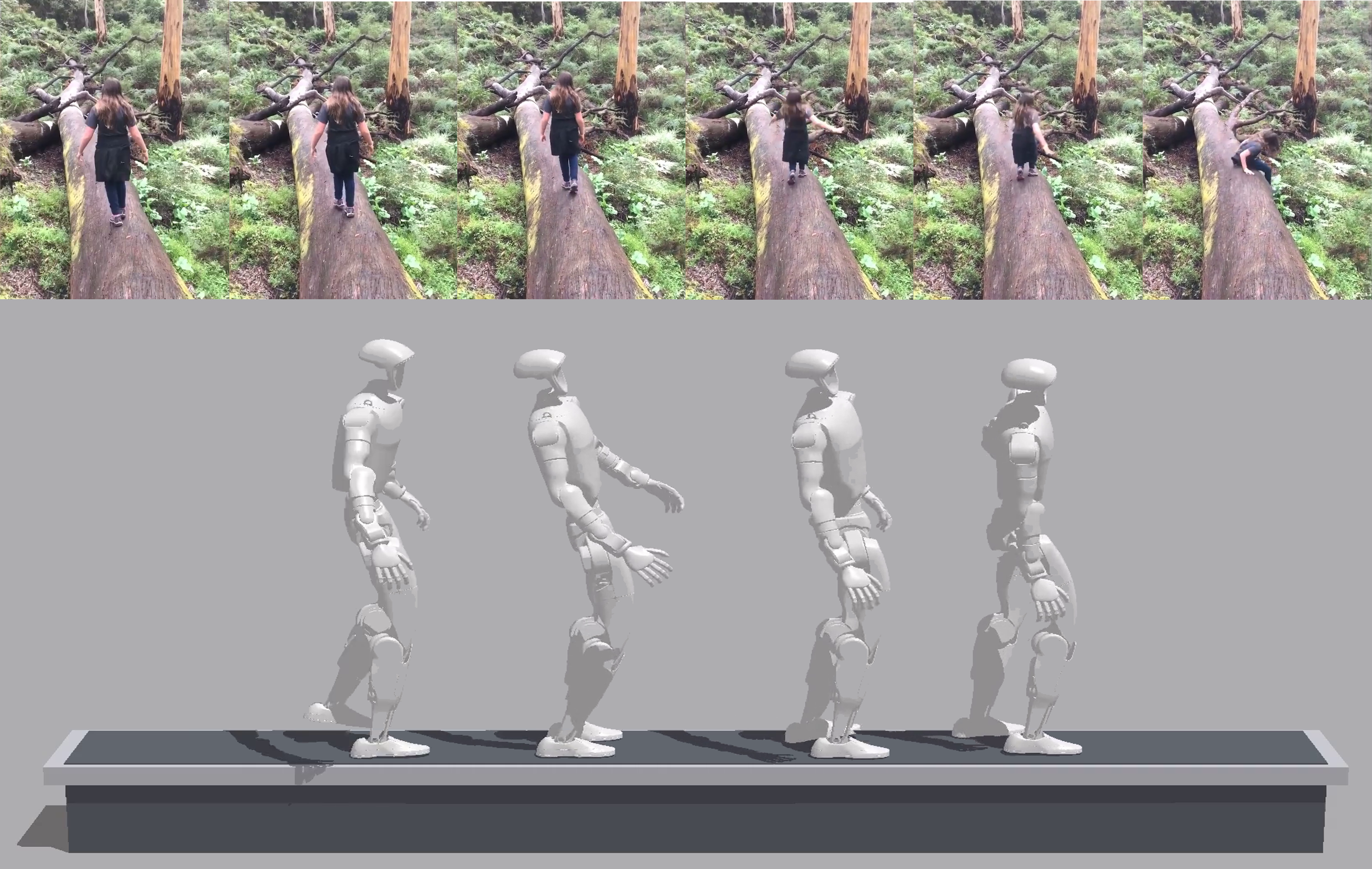}\\[0.4em]
  \includegraphics[width=0.97\linewidth]{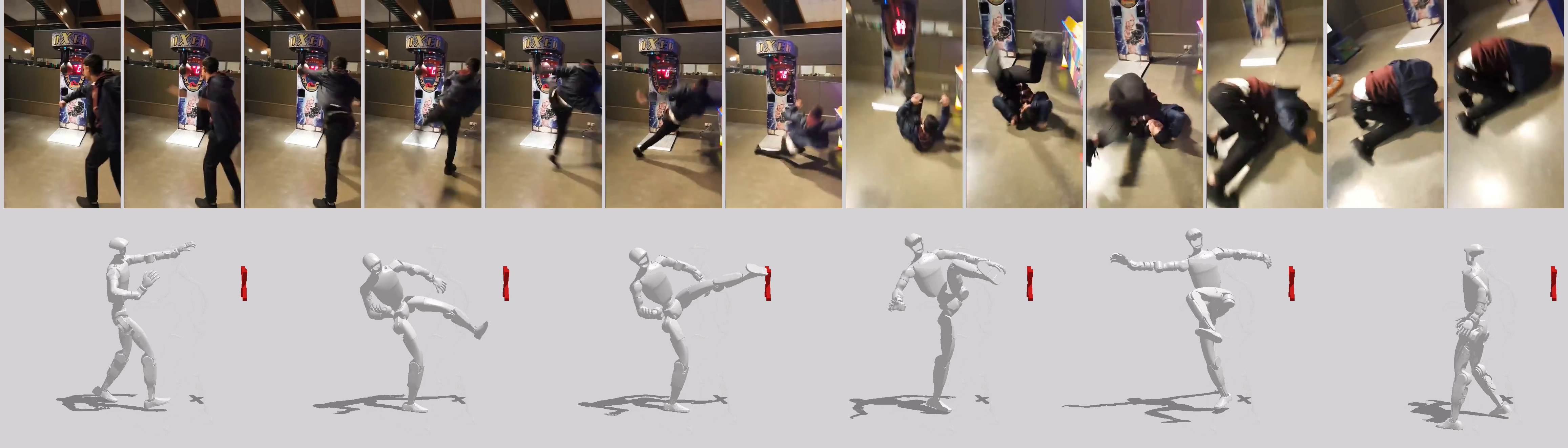}\\[0.4em]
  \includegraphics[width=0.97\linewidth]{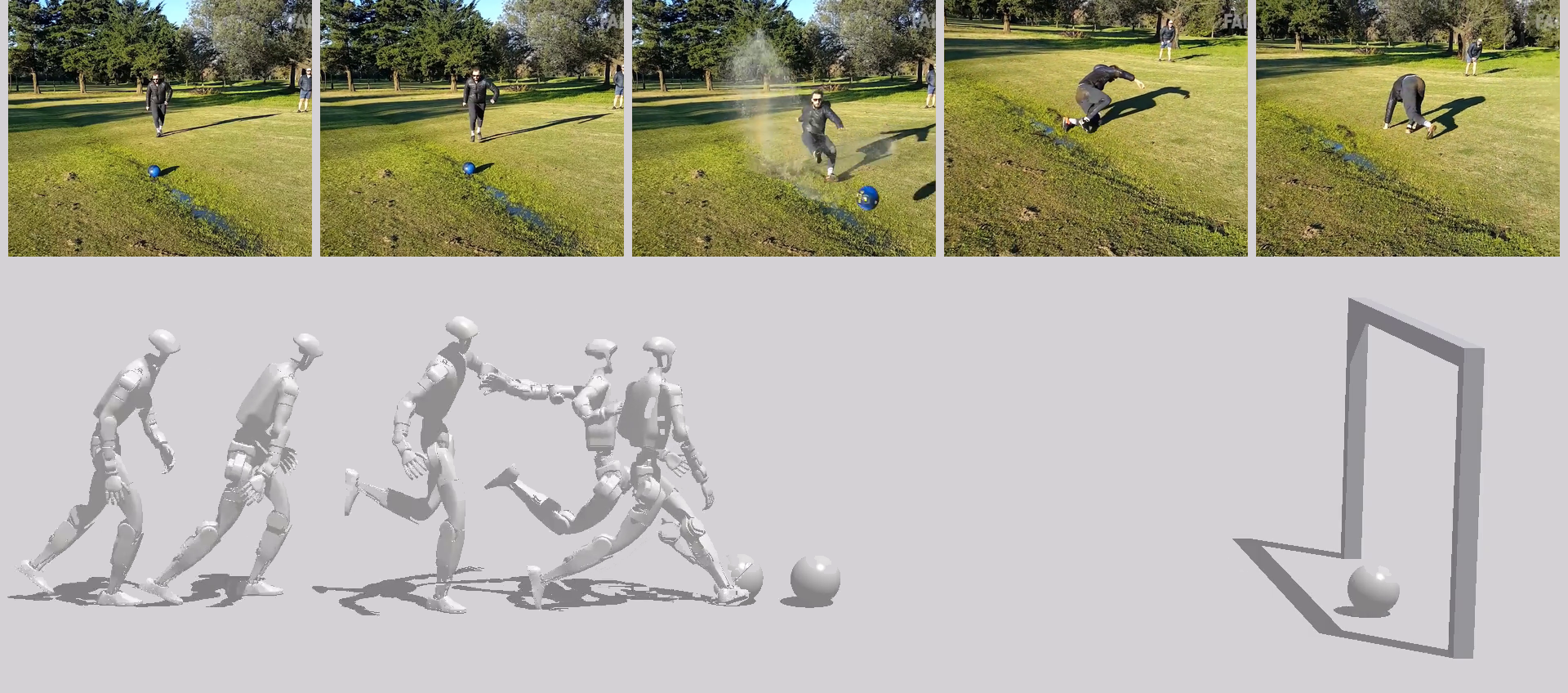}
  \caption{Failed human demonstrations (top) and corresponding policy rollouts (bottom) for log walking, target kicking, and football kicking, ordered from top to bottom. Frames progress from left to right. Best visualized in zoomed view.}
  \label{fig:policy_rollouts}
\end{figure}

\begin{table}[t]
  \centering
  \caption{Task success criteria.}
  \label{tab:success_criteria}
  \setlength{\tabcolsep}{2pt}
  \renewcommand{\arraystretch}{1.2}
  \scriptsize
  \begin{tabular}{@{}
      >{\raggedright\arraybackslash}p{0.155\columnwidth}
      >{\raggedright\arraybackslash}p{0.83\columnwidth}@{}}
    \toprule
    Task & Success criterion \\
    \midrule
    Kick target & Foot strikes the target at $\ge 2.0$\,m/s,
                  upright $\ge 0.80$, settled, held $2.0$\,s. \\
    Football    & Ball enters the goal region,
                  upright $\ge 0.80$, settled, held $2.0$\,s. \\
    Backflip    & Root becomes fully inverted, then both feet in the landing region and in
                  contact, upright $\ge 0.80$, settled,
                  held $2.0$\,s. \\
    Box jump    & Both feet on the box top,
                  upright $\ge 0.80$, settled, held $2.0$\,s. \\
    Handstand   & Both hands in the support region and in contact, feet at least
                  $0.40$\,m above the pelvis, held $0.5$\,s. \\
    Log walk    & Net displacement along the log $\ge 1.0$\,m without falling. \\
    \bottomrule
  \end{tabular}
\end{table}

\begin{figure*}[t]
  \centering
  \includegraphics[width=0.48\textwidth]{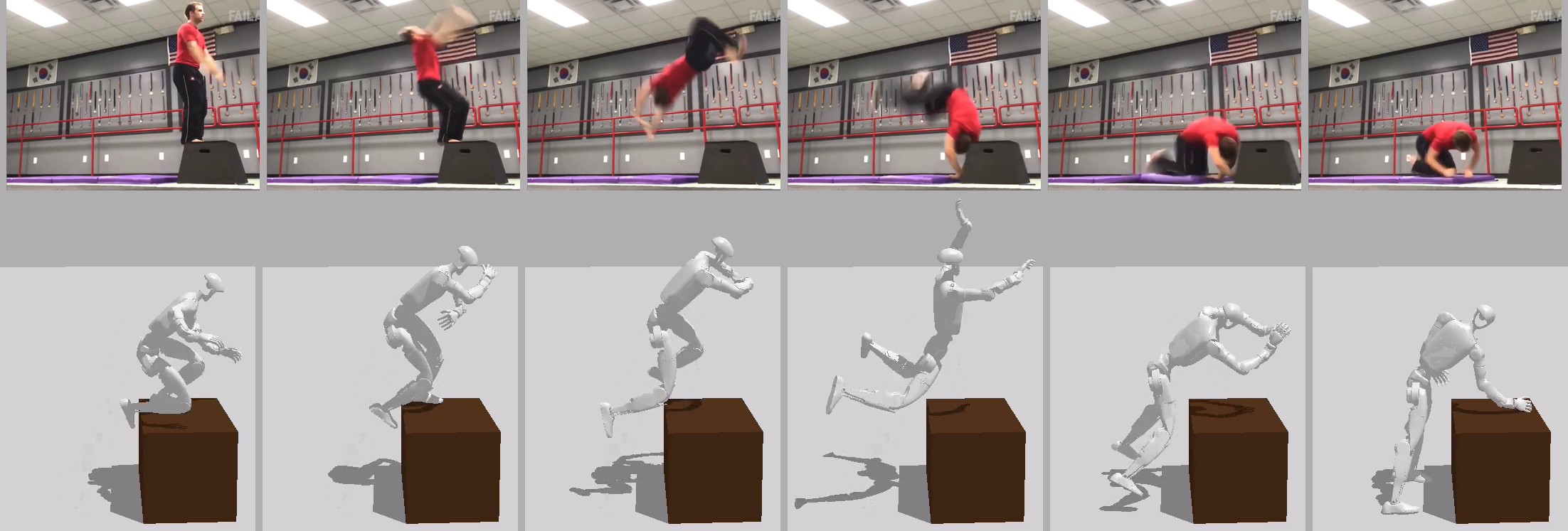}\hfill
  \includegraphics[width=0.48\textwidth]{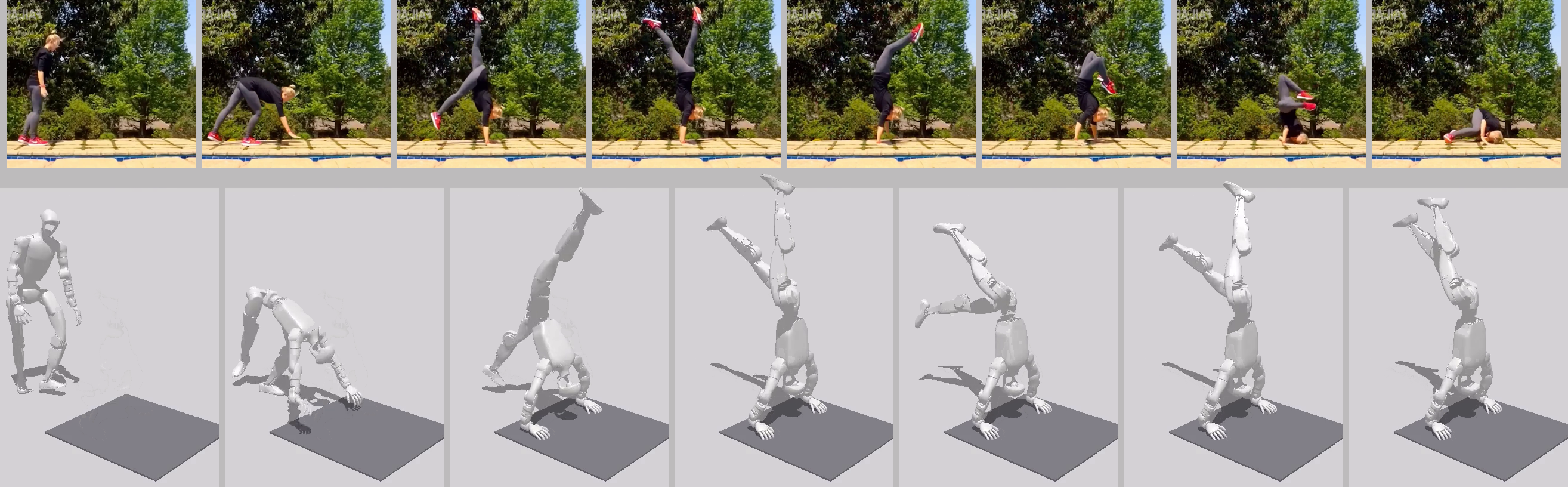}
  \caption{Failed human demonstrations (top) and corresponding policy rollouts (bottom) for backflip (left) and handstand (right).
  Frames progress from left to right. Best visualized in zoomed view.}
  \label{fig:diagnostic_rollouts}
\end{figure*}

\begin{table*}[t]
  \centering
  \caption{Comparison results for unified policy vs. task-reward-only policy.}
  \label{tab:prefix_reward_ablation}
  \setlength{\tabcolsep}{2.5pt}
  \renewcommand{\arraystretch}{1.18}
  \scriptsize
  \begin{tabular*}{0.94\textwidth}{@{\extracolsep{\fill}}l*{12}{c}@{}}
    \toprule
    \multirow{2}{*}{Policy}
      & \multicolumn{2}{c}{Kick target}
      & \multicolumn{2}{c}{Football}
      & \multicolumn{2}{c}{Backflip}
      & \multicolumn{2}{c}{Box jump}
      & \multicolumn{2}{c}{Handstand}
      & \multicolumn{2}{c}{Log walk} \\
    \cmidrule(lr){2-3}\cmidrule(lr){4-5}\cmidrule(lr){6-7}
    \cmidrule(lr){8-9}\cmidrule(lr){10-11}\cmidrule(lr){12-13}
      & \shortstack{Survival\\(\%)} & \shortstack{Success\\(\%)}
      & \shortstack{Survival\\(\%)} & \shortstack{Success\\(\%)}
      & \shortstack{Survival\\(\%)} & \shortstack{Success\\(\%)}
      & \shortstack{Survival\\(\%)} & \shortstack{Success\\(\%)}
      & \shortstack{Survival\\(\%)} & \shortstack{Success\\(\%)}
      & \shortstack{Survival\\(\%)} & \shortstack{Success\\(\%)} \\
    \midrule
    Task reward only
      & 100.0 & 0.0
      & 100.0 & 0.0
      & 100.0 & 0.0
      & 94.6  & 0.0
      & 100.0 & 0.0
      & 100.0 & 0.0 \\
    Unified policy (ours)
      & 100.0 & \textbf{99.3}
      & 100.0 & \textbf{100.0}
      & 99.7  & \textbf{41.7}
      & 98.6  & \textbf{75.8}
      & 100.0 & \textbf{56.3}
      & 100.0 & \textbf{100.0} \\
    \bottomrule
  \end{tabular*}
\end{table*}

\begin{figure}[t]
  \centering
  \includegraphics[width=\linewidth]{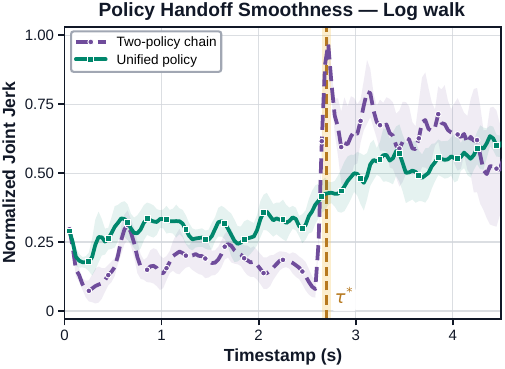}
  \caption{Normalized joint jerk around reference release on log walking.}
  \label{fig:jerk_ablation}
\end{figure}


\section{Experiments}
\label{sec:experiments}


\subsection{Experimental Setup}

\paragraph{Tasks and simulation}
Following the assumptions described in Sec.~\ref{sec:intro}, we select six failed human videos 
from the Oops! dataset~\cite{epstein2020oops}, including box jumping, log walking, target kicking, 
kicking a football into a goal, backflip, and handstand. 
Combinely, these six tasks cover impulsive whole-body motion, sustained narrow-support control, 
and object contact. 
All experiments use a 29-DoF Unitree G1 humanoid in Isaac Gym~\cite{makoviychuk2021isaacgym}. 
The policy outputs normalized joint-position targets at 30\,Hz, 
and training uses 4,096 parallel environments. 
We use GPT-5 (\texttt{gpt-5}) as the VLM in all experiments~\cite{openai2026gpt5}. 
To avoid evaluation bias, we define task success criteria manually instead of relying on the VLM-generated success criteria (see Table~\ref{tab:success_criteria}).

\paragraph{Policy and optimization}
Our policy architecture follows the tokenized Transformer design used for physical human-scene interaction in TokenHSI~\cite{pan2025tokenhsi}. Separate MLP encoders map proprioception, task, and reference motion states to 64-dimensional tokens. A four-layer, two-head Transformer encoder with a 512-dimensional feed-forward layer processes the three tokens, followed by a $[1024,512]$ action head.

We optimize all policies with PPO~\cite{schulman2017ppo}, using a 32-step rollout horizon, discount $\gamma=0.99$, GAE parameter $0.95$, clipping ratio $0.2$, and learning rate $2\times10^{-5}$. Each update uses 6-epochs and 8-minibatches. The prefix tracker is trained for 1,500 additional iterations and fine-tuned for 800 iterations from the initial reference frame. Each valid task reward candidate receives the same 300-iteration training budget, and the candidate with the highest terminal success is further fine-tuned for 1,500 iterations to train a unified policy.

Table~\ref{tab:observation_space} summarizes the policy observation. The task block contains the object, goal, and scene quantities required by each task, so its dimension $d_{\mathrm{task}}$ varies across tasks. The 18-dimensional reference block is masked-out to zero at $\tau$.

\subsection{Evaluation Results}
\label{sec:eval_results}

Figures~\ref{fig:policy_rollouts} and~\ref{fig:diagnostic_rollouts} show representative policy rollouts. These figures illustrates that the usable prefix supplies task specific preparation, while the unified policy replaces the failed continuation with motion directed toward the intended outcome. Fig.~\ref{fig:ponr_curves} evaluates the corresponding per-task policies over fixed release times. The VLM-proposed PoNR window is the visual training prior introduced in Sec.~\ref{sec:method}, whereas the measured PoNR $\tau^*$ is obtained from the trained policy's curve. Since the release time $\tau$ is provided to the policy as an observation, task-success outside the sampled window $PoNR_w$ shows the temporal generalization of the policy rather than dependence on one hand-selected release time.

\paragraph{Kick target}
The recovered motion provide the intial step-in, support-leg alignment, and leg swing, whereas 
the reaching the target, balancing on the support leg, and recovering from the kick are learned from the task reward. 
PoNR curve shows that the success remains high beyond the $PoNR_w$, indicating that the policy learned to recover from failling-states. It fail to recover once the reference extends into the purely unusable trajectory motion, placing the measured PoNR $\tau^*$ after the visual window but before the trajectory becomes harmful.

\paragraph{Football}
The early non-monotonic region indicates sensitivity to the state at which guidance is removed. 
Once the reference reaches a useful approach and leg-swing state, it successfully performs the task. 
Furthermore, Table~\ref{tab:prefix_reward_ablation} demonstrates that the unified policy outperforms the task-reward-only policy in terms of success rate,
showing the importance of the motion prefix for task completion. 

\paragraph{Backflip}
The backflip task exposes a limitation of the selected reward rather than the success measure.
Inversion contributes only 0.05 to the reward (Table~\ref{tab:selected_rewards}), whereas the success criterion requires a full inversion before landing (Table~\ref{tab:success_criteria}).
When released early, the policy jumps backward and lands without performing a full backflip, achieving high survival but low task success.
The PoNR curve reflects this behavior, as the success rate increases once the reference motion provides the inverted state. After this point, the policy only needs to land and settle instead of learning the full backflip.

\paragraph{Box jump}
The policy rollout frames for box jump task is shown in Fig.~\ref{fig:teaser}. 
The approach and crouched loading motion prepare the humanoid for takeoff. 
The PoNR curve explains the behavior of intermediate releases, which expose a useful state from which the policy completes the jump and landing, while later releases sharply reduce success as the reference enters the failed continuation.

\paragraph{Handstand}
The handstand task clearly separates visual failure from dynamic usefulness. 
The VLM places the release window near the early appearance of failure, approximately $2.5$--$2.8$\,s, 
whereas the greatest task success occurs near $6.7$\,s. 
This phenomenon arises because the later reference motion already produces an inverted hand-supported state, 
leaving the policy only to stabilize the pose for the remaining hold duration.

\paragraph{Log walk}
Instead of simple walking action, we uplifted the task difficulty to learn to maintain balance on the narrow support. 
In this log walk task, the prefix trajectory provides the approach and initial balance-oriented steps, and the task completion reward requires
to maintain balance and walk for $\ge 1.0$\,m to achieve success.

Overall, the VLM-proposed window can only be treated as a training prior rather than the final PoNR $\tau^*$. The useful boundary depends on both the task and the trained policy and is therefore determined from the release-time curve.


\subsection{Ablation Studies}
\label{sec:exp_unified_ablation}

Existing video-to-humanoid methods require a target demonstration, 
while prior learning from failure methods use different inputs or 
learning objectives (see Table~\ref{tab:prior_work_comparison}). 
Therefore, no prior method directly matches our 
single failed video setting. 
We evaluate the two main design choices of our method 
through ablation studies.

\paragraph{Prefix guidance vs. task reward only}
We perform an ablation to evaluate the effectiveness of prefix guidance.
We compare our unified policy training with task-reward-only policies by removing the prefix trajectory guidance and training only with the selected task rewards (see Table~\ref{tab:prefix_reward_ablation}).
The reward-only policies remain stable and achieve high survival success, but fail to satisfy the terminal success criteria for all six tasks.
In contrast, the unified policies produce non-zero success on every task, showing that the prefix provides task-specific approach and interaction states that are difficult to discover from the task reward alone.

\paragraph{Unified policy vs. policy chaining}
This ablation evaluates why \methodName~uses one policy before and after $\tau$. 
The natural alternative to our approach is a two-policy chaining mechanism~\cite{lee2022tstar, uchendu2023jsrl}, 
where a prefix policy tracks the reference until $\tau$ and a separate completion policy trained with the task reward.
We use log walking as a test scenario for this ablation because the narrow-support locomotion 
makes a control discontinuity immediately visible. 
We align rollouts at $\tau^*$ and measure the normalized joint jerk over the complete episode. As shown in Fig.~\ref{fig:jerk_ablation}, 
the two-policy chain rises from around 0.17 before release to 0.90 at the switch, 
whereas the unified policy changes from 0.30 to 0.42. 
The unified policy therefore has 2.1$\times$ lower jerk at release $\tau$. The result specifically shows that retaining the same controller avoids the sharp discontinuity caused by switching between independently trained policies.


\section{Discussion and Future Work}
\label{sec:discussion}

Our results show that a failed video can provide useful supervision when its motion prefix and intended outcome are used in tandem. 
The release-time curves further show that visual failure does not always coincide with the point at which the recovered 
motion becomes dynamically harmful, supporting policy-conditioned evaluation of the PoNR. 
These findings make learning from failed demonstrations a promising direction, 
but the current approach is limited to one humanoid embodiment and non-dexterious humanoid-object interaction tasks. 

Future work will extend motion reconstruction and retargeting to dexterous 
hand motion, enabling robotic manipulation tasks that require precise hand-object interaction. We also plan to move beyond a single humanoid and study failed demonstrations in multi-humanoid collaborative tasks.

\section{Conclusion}
\label{sec:conclusion}

We present \methodName, a framework that learns humanoid skills from a 
single failed human video without requiring a successful task motion 
demonstration. By treating the recovered prefix as partial supervision, 
inferring the intended outcome, and training one release time conditioned policy, 
\methodName~retains useful preparation while 
replacing the failed continuation with task completion-related behavior. 
The experiments demonstrate the importance of prefix guidance, and show that a unified policy reduces the release discontinuity introduced by policy chaining.
The results show that failed human videos can provide useful information for humanoid skill learning and establish a new setting for future research in learning from unsuccessful demonstrations.



\bibliographystyle{IEEEtran}
\bibliography{references}

\begin{thebibliography}{10}
\providecommand{\url}[1]{#1}
\csname url@samestyle\endcsname
\providecommand{\newblock}{\relax}
\providecommand{\bibinfo}[2]{#2}
\providecommand{\BIBentrySTDinterwordspacing}{\spaceskip=0pt\relax}
\providecommand{\BIBentryALTinterwordstretchfactor}{4}
\providecommand{\BIBentryALTinterwordspacing}{\spaceskip=\fontdimen2\font plus
\BIBentryALTinterwordstretchfactor\fontdimen3\font minus \fontdimen4\font\relax}
\providecommand{\BIBforeignlanguage}[2]{{%
\expandafter\ifx\csname l@#1\endcsname\relax
\typeout{** WARNING: IEEEtran.bst: No hyphenation pattern has been}%
\typeout{** loaded for the language `#1'. Using the pattern for}%
\typeout{** the default language instead.}%
\else
\language=\csname l@#1\endcsname
\fi
#2}}
\providecommand{\BIBdecl}{\relax}
\BIBdecl

\bibitem{allshire2025videomimic}
\BIBentryALTinterwordspacing
A.~Allshire, H.~Choi, J.~Zhang, D.~McAllister, A.~Zhang, C.~M. Kim, T.~Darrell, P.~Abbeel, J.~Malik, and A.~Kanazawa, ``Visual imitation enables contextual humanoid control,'' in \emph{Proceedings of the 9th Conference on Robot Learning}, ser. Proceedings of Machine Learning Research, vol. 305.\hskip 1em plus 0.5em minus 0.4em\relax PMLR, 2025, pp. 794--815, best Student Paper Award. [Online]. Available: \url{https://proceedings.mlr.press/v305/allshire25a.html}
\BIBentrySTDinterwordspacing

\bibitem{zhang2026meshmimic}
Q.~Zhang, J.~Ma, P.~Liu, S.~Shi, Z.~Su, Z.~Wang, J.~Sun, W.~Cui, J.~Yu, G.~Han \emph{et~al.}, ``Meshmimic: Geometry-aware humanoid motion learning through 3d scene reconstruction,'' \emph{arXiv preprint arXiv:2602.15733}, 2026.

\bibitem{weng2025hdmi}
H.~Weng, Y.~Li, N.~Sobanbabu, Z.~Wang, Z.~Luo, T.~He, D.~Ramanan, and G.~Shi, ``Hdmi: Learning interactive humanoid whole-body control from human videos,'' \emph{arXiv preprint arXiv:2509.16757}, 2025.

\bibitem{okami2024}
J.~Li, Y.~Zhu, Y.~Xie, Z.~Jiang, M.~Seo, G.~Pavlakos, and Y.~Zhu, ``Okami: Teaching humanoid robots manipulation skills through single video imitation,'' in \emph{8th Annual Conference on Robot Learning (CoRL)}, 2024.

\bibitem{zhu2026vision}
Y.~Zhu, A.~Lim, P.~Stone, and Y.~Zhu, ``Vision-based manipulation from single human video with open-world object graphs,'' \emph{Autonomous Robots}, vol.~50, no.~27, 2026.

\bibitem{gupta2026lucid}
H.~Gupta, G.~Shi, and W.~Yuan, ``Lucid: Learning embodiment-agnostic intent models from unstructured human videos for scalable dexterous robot skill acquisition,'' \emph{arXiv preprint arXiv:2606.11628}, 2026.

\bibitem{grollman2011donut}
D.~H. Grollman and A.~Billard, ``Donut as i do: Learning from failed demonstrations,'' in \emph{ICRA}, 2011.

\bibitem{wang2024}
Y.~Wang, Z.~Sun, J.~Zhang, Z.~Xian, E.~Biyik, D.~Held, and Z.~Erickson, ``Rl-vlm-f: Reinforcement learning from vision language foundation model feedback,'' in \emph{Proceedings of the 41st International Conference on Machine Learning}, 2024.

\bibitem{epstein2021learning}
D.~Epstein and C.~Vondrick, ``Learning goals from failure,'' in \emph{2021 IEEE/CVF Conference on Computer Vision and Pattern Recognition (CVPR)}.\hskip 1em plus 0.5em minus 0.4em\relax IEEE, 2021, pp. 11\,189--11\,199.

\bibitem{peng2018sfv}
X.~B. Peng, A.~Kanazawa, J.~Malik, P.~Abbeel, and S.~Levine, ``Sfv: reinforcement learning of physical skills from videos,'' in \emph{ACM Trans. Graph. (SIGGRAPH Asia)}, 2018.

\bibitem{luo2023perpetual}
Z.~Luo, J.~Cao, K.~Kitani, W.~Xu \emph{et~al.}, ``Perpetual humanoid control for real-time simulated avatars,'' in \emph{Proceedings of the IEEE/CVF International Conference on Computer Vision}, 2023, pp. 10\,895--10\,904.

\bibitem{he2024h2o}
T.~He, Z.~Luo, W.~Xiao, C.~Zhang, K.~Kitani, C.~Liu, and G.~Shi, ``Learning human-to-humanoid real-time whole-body teleoperation,'' in \emph{2024 IEEE/RSJ International Conference on Intelligent Robots and Systems (IROS)}.\hskip 1em plus 0.5em minus 0.4em\relax IEEE, 2024, pp. 8944--8951.

\bibitem{epstein2020oops}
D.~Epstein, B.~Chen, and C.~Vondrick, ``Oops! predicting unintentional action in video,'' in \emph{2020 IEEE/CVF Conference on Computer Vision and Pattern Recognition (CVPR)}.\hskip 1em plus 0.5em minus 0.4em\relax IEEE, 2020, pp. 916--926.

\bibitem{mahmood2019amass}
N.~Mahmood, N.~Ghorbani, N.~F. Troje, G.~Pons-Moll, and M.~J. Black, ``Amass: Archive of motion capture as surface shapes,'' in \emph{Proceedings of the IEEE/CVF international conference on computer vision}, 2019, pp. 5442--5451.

\bibitem{peng2018deepmimic}
\BIBentryALTinterwordspacing
X.~B. Peng, P.~Abbeel, S.~Levine, and M.~van~de Panne, ``{DeepMimic}: Example-guided deep reinforcement learning of physics-based character skills,'' \emph{ACM Transactions on Graphics}, vol.~37, no.~4, pp. 1--14, 2018. [Online]. Available: \url{https://doi.org/10.1145/3197517.3201311}
\BIBentrySTDinterwordspacing

\bibitem{peng2021amp}
X.~B. Peng, Z.~Ma, P.~Abbeel, S.~Levine, and A.~Kanazawa, ``Amp: adversarial motion priors for stylized physics-based character control,'' in \emph{ACM Trans. Graph. (SIGGRAPH)}, 2021.

\bibitem{peng2022ase}
X.~B. Peng, Y.~Guo, L.~Halper, S.~Levine, and S.~Fidler, ``Ase: Large-scale reusable adversarial skill embeddings for physically simulated characters,'' \emph{ACM Transactions On Graphics (TOG)}, vol.~41, no.~4, pp. 1--17, 2022.

\bibitem{tessler2023calm}
C.~Tessler, Y.~Kasten, Y.~Guo, S.~Mannor, G.~Chechik, and X.~B. Peng, ``Calm: Conditional adversarial latent models for directable virtual characters,'' in \emph{ACM SIGGRAPH 2023 conference proceedings}, 2023, pp. 1--9.

\bibitem{shen2024gvhmr}
\BIBentryALTinterwordspacing
Z.~Shen, H.~Pi, Y.~Xia, Z.~Cen, S.~Peng, Z.~Hu, H.~Bao, R.~Hu, and X.~Zhou, ``World-grounded human motion recovery via gravity-view coordinates,'' in \emph{SIGGRAPH Asia 2024 Conference Papers}.\hskip 1em plus 0.5em minus 0.4em\relax ACM, 2024, pp. 1--11. [Online]. Available: \url{https://doi.org/10.1145/3680528.3687565}
\BIBentrySTDinterwordspacing

\bibitem{li2025robomirror}
Z.~Li, C.~Chi, B.~Zhu, Y.~Wei, S.~Bai, Y.~Ji, Y.~Peng, T.~Huang, P.~Wang, Z.~Wang \emph{et~al.}, ``Robomirror: Understand before you imitate for video to humanoid locomotion,'' \emph{arXiv preprint arXiv:2512.23649}, 2025.

\bibitem{wu2019imperfect}
\BIBentryALTinterwordspacing
Y.-H. Wu, N.~Charoenphakdee, H.~Bao, V.~Tangkaratt, and M.~Sugiyama, ``Imitation learning from imperfect demonstration,'' in \emph{ICML}, 2019. [Online]. Available: \url{https://arxiv.org/abs/1901.09387}
\BIBentrySTDinterwordspacing

\bibitem{brown2019trex}
\BIBentryALTinterwordspacing
D.~S. Brown, W.~Goo, P.~Nagarajan, and S.~Niekum, ``Extrapolating beyond suboptimal demonstrations via inverse reinforcement learning from observations,'' in \emph{ICML}, 2019. [Online]. Available: \url{https://arxiv.org/abs/1904.06387}
\BIBentrySTDinterwordspacing

\bibitem{yu2023language}
\BIBentryALTinterwordspacing
W.~Yu, N.~Gileadi, C.~Fu, S.~Kirmani, K.-H. Lee, M.~Gonzalez~Arenas, H.-T.~L. Chiang, T.~Erez, L.~Hasenclever, J.~Humplik, B.~Ichter, T.~Xiao, P.~Xu, A.~Zeng, T.~Zhang, N.~Heess, D.~Sadigh, J.~Tan, Y.~Tassa, and F.~Xia, ``Language to rewards for robotic skill synthesis,'' in \emph{Proceedings of the 7th Conference on Robot Learning}, ser. Proceedings of Machine Learning Research, vol. 229.\hskip 1em plus 0.5em minus 0.4em\relax PMLR, 2023. [Online]. Available: \url{https://proceedings.mlr.press/v229/yu23a.html}
\BIBentrySTDinterwordspacing

\bibitem{xie2023text2reward}
\BIBentryALTinterwordspacing
T.~Xie, S.~Zhao, C.~H. Wu, Y.~Liu, Q.~Luo, V.~Zhong, Y.~Yang, and T.~Yu, ``Text2reward: Reward shaping with language models for reinforcement learning,'' in \emph{ICLR}, 2023. [Online]. Available: \url{https://arxiv.org/abs/2309.11489}
\BIBentrySTDinterwordspacing

\bibitem{ma2024eureka}
\BIBentryALTinterwordspacing
Y.~J. Ma, W.~Liang, G.~Wang, D.-A. Huang, O.~Bastani, D.~Jayaraman, Y.~Zhu, L.~Fan, and A.~Anandkumar, ``{Eureka}: Human-level reward design via coding large language models,'' in \emph{The Twelfth International Conference on Learning Representations (ICLR)}, 2024. [Online]. Available: \url{https://openreview.net/forum?id=HfVc7hq3ag}
\BIBentrySTDinterwordspacing

\bibitem{ma2024dreureka}
\BIBentryALTinterwordspacing
Y.~J. Ma, W.~Liang, H.-J. Wang, S.~Wang, Y.~Zhu, L.~Fan, O.~Bastani, and D.~Jayaraman, ``Dreureka: Language model guided sim-to-real transfer,'' in \emph{RSS}, 2024. [Online]. Available: \url{https://arxiv.org/abs/2406.01967}
\BIBentrySTDinterwordspacing

\bibitem{zeng2024video2reward}
\BIBentryALTinterwordspacing
R.~Zeng, D.~Zhou, Q.~Liang, J.~Liu, H.~Li, C.~Huang, J.~Li, X.~Hu, and F.~Sun, ``{Video2Reward}: Generating reward function from videos for legged robot behavior learning,'' in \emph{ECAI 2024 -- 27th European Conference on Artificial Intelligence}, ser. Frontiers in Artificial Intelligence and Applications.\hskip 1em plus 0.5em minus 0.4em\relax IOS Press, 2024, pp. 4369--4376. [Online]. Available: \url{https://doi.org/10.3233/FAIA241014}
\BIBentrySTDinterwordspacing

\bibitem{rocamonde2023vlmrm}
\BIBentryALTinterwordspacing
J.~Rocamonde, V.~Montesinos, E.~Nava, E.~Perez, and D.~Lindner, ``Vision-language models are zero-shot reward models for reinforcement learning,'' in \emph{ICLR}, 2023. [Online]. Available: \url{https://arxiv.org/abs/2310.12921}
\BIBentrySTDinterwordspacing

\bibitem{makoviychuk2021isaacgym}
\BIBentryALTinterwordspacing
V.~Makoviychuk, L.~Wawrzyniak, Y.~Guo, M.~Lu, K.~Storey, M.~Macklin, D.~Hoeller, N.~Rudin, A.~Allshire, A.~Handa, and G.~State, ``{Isaac Gym}: High performance {GPU}-based physics simulation for robot learning,'' in \emph{Proceedings of the Neural Information Processing Systems Track on Datasets and Benchmarks (NeurIPS)}, 2021. [Online]. Available: \url{https://datasets-benchmarks-proceedings.neurips.cc/paper/2021/hash/28dd2c7955ce926456240b2ff0100bde-Abstract-round2.html}
\BIBentrySTDinterwordspacing

\bibitem{openai2026gpt5}
\BIBentryALTinterwordspacing
{OpenAI}, ``Openai gpt-5 system card,'' \emph{CoRR}, vol. abs/2601.03267, 2026. [Online]. Available: \url{https://doi.org/10.48550/arXiv.2601.03267}
\BIBentrySTDinterwordspacing

\bibitem{pan2025tokenhsi}
\BIBentryALTinterwordspacing
L.~Pan, Z.~Yang, Z.~Dou, W.~Wang, B.~Huang, B.~Dai, T.~Komura, and J.~Wang, ``{TokenHSI}: Unified synthesis of physical human--scene interactions through task tokenization,'' in \emph{Proceedings of the IEEE/CVF Conference on Computer Vision and Pattern Recognition (CVPR)}, June 2025, pp. 5379--5391. [Online]. Available: \url{https://arxiv.org/abs/2503.19901}
\BIBentrySTDinterwordspacing

\bibitem{schulman2017ppo}
\BIBentryALTinterwordspacing
J.~Schulman, F.~Wolski, P.~Dhariwal, A.~Radford, and O.~Klimov, ``Proximal policy optimization algorithms,'' 2017. [Online]. Available: \url{https://arxiv.org/abs/1707.06347}
\BIBentrySTDinterwordspacing

\bibitem{lee2022tstar}
\BIBentryALTinterwordspacing
Y.~Lee, J.~J. Lim, A.~Anandkumar, and Y.~Zhu, ``Adversarial skill chaining for long-horizon robot manipulation via terminal state regularization,'' in \emph{Proceedings of the 5th Conference on Robot Learning}, ser. Proceedings of Machine Learning Research, vol. 164.\hskip 1em plus 0.5em minus 0.4em\relax PMLR, 2022, pp. 406--416. [Online]. Available: \url{https://proceedings.mlr.press/v164/lee22a.html}
\BIBentrySTDinterwordspacing

\bibitem{uchendu2023jsrl}
\BIBentryALTinterwordspacing
I.~Uchendu, T.~Xiao, Y.~Lu, B.~Zhu, M.~Yan, J.~Simon, M.~Bennice, C.~Fu, C.~Ma, J.~Jiao, S.~Levine, and K.~Hausman, ``Jump-start reinforcement learning,'' in \emph{Proceedings of the 40th International Conference on Machine Learning}, ser. Proceedings of Machine Learning Research, vol. 202.\hskip 1em plus 0.5em minus 0.4em\relax PMLR, 2023, pp. 34\,556--34\,583. [Online]. Available: \url{https://proceedings.mlr.press/v202/uchendu23a.html}
\BIBentrySTDinterwordspacing

\end{thebibliography}

\end{document}